\documentclass[letterpaper]{article}
\usepackage{ijcai17}
\usepackage{helvet}
\usepackage{courier}
\usepackage{times}
\usepackage{latexsym}
\usepackage{graphicx}
\usepackage{color}
\usepackage{multirow}
\usepackage{amsmath}
\usepackage{amssymb}
\usepackage[usenames,dvipsnames]{xcolor}
\usepackage[ruled,vlined,linesnumbered]{algorithm2e}
\usepackage{float}
\usepackage{tabularx}
\usepackage{multicol}

\newtheorem{definition}{Definition}

\newcommand{\comment}[1]{}

\newtheorem{corollary}{Corollary}

\makeindex

\begin{document}

\title{Modifying Optimal SAT-based Approach to Multi-agent Path-finding Problem to Suboptimal Variants}

\author{Pavel Surynek\\
National Institute of Advanced Industrial\\
Science and Technology (AIST)\\
Tokyo 135-0064, Japan\\
pavel.surynek@aist.go.jp
\And
Ariel Felner~~~~Roni Stern\\
Ben Gurion University\\
Beer-Sheva, Israel 84105\\
felner,sternron@bgu.ac.il
\And
Eli Boyarski\\
Bar-Ilan University\\
Ramat-Gan, Israel\\
eli.boyarski@gmail.com
}

\maketitle
\begin{small}
\begin{abstract}
In {\em multi-agent path finding} (MAPF) the task is to find
non-conflicting paths for multiple agents. In this paper we focus on finding
suboptimal solutions for MAPF for the {\em sum-of-costs} variant.
Recently, a SAT-based approached was developed to solve this problem and proved
beneficial in many cases when compared to other search-based solvers. In this
paper, we present SAT-based unbounded- and bounded-suboptimal algorithms and
compare them to relevant algorithms. Experimental results show that in many
case the SAT-based solver significantly outperforms the search-based solvers.

\end{abstract}
\end{small}

\section{Introduction}
\noindent \noindent


The {\em multi-agent path finding} (MAPF) problem consists  a graph, $G=(V,E)$
and a set $A=\{a_1, a_2,\dots a_k\}$ of $k$ agents. Time is discretized into
time steps. The arrangement of agents at time-step $t$ is denoted as
$\alpha_t$. Each agent $a_i$ has a start position $\alpha_0(a_i) \in V$ and a
goal position $\alpha_+(a_i) \in V$.  At each time step an agent can either
{\em move} to an adjacent location or {\em wait} in its current location.
The task is to find a sequence of move/wait actions for each agent $a_i$, moving it from
$\alpha_0(a_i)$ to $\alpha_+(a_i)$ such that agents do not {\em conflict},
i.e., do not occupy the same location at the same time. Formally, an MAPF
instance is a tuple $\Sigma=(G=(V,E),A,\alpha_0,\alpha_+)$. A \textit{solution}
for $\Sigma$ is a sequence of arrangements
$\mathcal{S}(\Sigma)=[\alpha_0,\alpha_1,...,\alpha_{\mu}]$ such that
 $\alpha_{t+1}$ results from valid movements from
$\alpha_{t}$ for $t=1,2,...,\mu-1$,
and $\alpha_{\mu}=\alpha_+$.

MAPF has practical applications in video games, traffic control, robotics etc.
(see~\cite{CBSJUR} for a survey). The scope of this paper is limited to the
setting of {\em fully cooperative} agents that are centrally controlled. MAPF
is usually solved aiming to minimize one of the two commonly-used global
cumulative cost functions: {\bf (1)  Sum-of-costs} (denoted $\xi$) is the
summation, over all agents, of the number of time steps required to reach the
goal
location~\cite{dresner2008aMultiagent,standley2010finding,DBLP:journals/ai/SharonSGF13,CBSJUR}.
Formally, $\xi = \sum_{i=1}^k{\xi(a_i)}$, where $\xi(a_i)$ is an
\textit{individual path cost} of agent $a_i$. {\bf (2) Makespan:} (denoted
$\mu$) is the time until the last agent reaches its destination (i.e.,
the maximum of the individual
costs)~\cite{DBLP:conf/aaai/Surynek10,DBLP:conf/ictai/Surynek14,DBLP:conf/ijcai/Surynek15}.



Optimal solvers for MAPF can be divided to two classes.
{\bf (1) Search-based solvers.} These algorithms consider MAPF as a graph search problem. Some of these algorithm are variants of the A* algorithm that search in a global {\em search space} -- all different ways to place $k$ agents into $V$ vertices, one agent per vertex~\cite{standley2010finding,DBLP:journals/ai/WagnerC15}.
Others algorithms such as \textsc{Icts}~\cite{DBLP:journals/ai/SharonSGF13} and \textsc{Cbs}~\cite{CBSJUR,DBLP:conf/ijcai/BoyarskiFSSTBS15} search different search spaces and employ novel (non-A*) search tree. All these search-based solvers were originally designed for the {\em sum-of-costs} MAPF variant.
But, with simple modifications, they can
be modified to work for the makespan variant. \noindent{\bf (2)
Reduction-based solvers}. By contrast, many recent optimal solvers
reduce MAPF to known problems such as
CSP~\cite{DBLP:conf/icra/Ryan10}, SAT~\cite{surynek2012towards},
Inductive Logic Programming~\cite{yu2013planning} and Answer Set
Programming~\cite{erdem2013general}. 
While most reduction-based solvers address the makespan variant, an optimal
reduction-based solver for the sum-of-costs variant was recently
introduced~\cite{surynek2016empirical}. In this paper we further widen this
direction and introduce {\bf SAT-based suboptimal solvers}.

Finding optimal solutions for both variants is
NP-Hard~\cite{DBLP:conf/aaai/YuL13,DBLP:conf/aaai/Surynek10}; as the
state-space grows exponentially with $k$ (\# of agents). Therefore, many
suboptimal solvers were developed. Some suboptimal solvers aim to to quickly
find paths for all agents while paying no attention to the quality of the
solution, i.e., how far it is from the optimal solution. We refer to such
algorithms as {\bf any solution} MAPF solvers. Many any solution MAPF solvers
were
proposed~\cite{DBLP:conf/icra/Ryan10,CohenUK15,DBLP:conf/aiide/Silver05,BoteaS15,sajid2012multi},
and there is even a polynomial time any solution MAPF solver. These algorithms
are usually used when $k$ is large and some of  them are not  complete.


In some cases, the user might ask for some guarantee on the
quality of the solution returned. A common type of such a requirement
is that the solution found is {\bf
bounded suboptimal}, that its cost is $\leq (1+\epsilon)
\times c_{opt}$ where $c_{opt}$ is the cost of the optimal solution and $\epsilon$ is a parameter that sets the desired amount of suboptimality - sometimes called the {\em error}. A solver that returns bounded-suboptimal solutions is referred to as a {\bf
bounded-suboptimal} algorithm or more specifically $(1+\epsilon)$-bounded suboptimal.

Despite the large number of papers devoted to optimal or to suboptimal
solutions, we are only aware of two approaches that provided bounded suboptimal
solutions \textsc{Ecbs}~\cite{barer2014suboptimal} and \textsc{Cbs} with
highways~\cite{CohenUK15}, both are modifications of the {\em conflict based
search} (\textsc{Cbs}) algorithm. In this paper we introduce two new SAT-based
solvers: u\textsc{Mdd-Sat}, an any solution MAPF solver, and e\textsc{Mdd-Sat}, a bounded-suboptimal MAPF solver. 
We experimentally compare our new SAT solvers with
relevant any solution or bounded-suboptimal algorithms and show that our SAT
solvers is comparable and sometimes outperform other algorithms in many circumstances.

\section{Background: Optimal SAT-based Solver}

Our suboptimal algorithms presented in this paper are based on a SAT-based
optimal MAPF (called \textsc{Mdd-Sat}) algorithm for the sum-of-costs variant which was
~\cite{SurynekFSB16}. The main idea in \textsc{Mdd-Sat} is to convert the optimization
problem (finding minimal sum-of-costs) to a sequence of decision problems -- is
there a solution of a given sum-of-costs $\xi$. A formula $\mathcal{F}_\xi$ has
been introduced such that $\mathcal{F}_\xi$ is satisfiable if and only if there
is a solution of sum-of-costs $\xi$. We now provide sufficient details about
$\mathcal{F}_\xi$ that are needed for the rest of this paper. More information
on this formula and its exact variables can be found in~\cite{SurynekFSB16}.

Let $\xi_0(a_i)$ be the cost of the shortest individual path for agent $a_i$
(ignoring collisions with the other agents), and let $\xi_0=\sum_{a_i\in A}
\xi_0(a_i)$. $\xi_0$ is called the {\em sum of individual costs}
(SIC)~\cite{DBLP:journals/ai/SharonSGF13}. It is a known admissible heuristic
for optimal sum-of-costs search algorithms, since it is a lower bound on the
minimal sum-of-costs. $\xi_0$ is calculated by relaxing the problem by omitting
the other agents, solving $k$ single-agent shortest path problem.  Similarly,
we define $\mu_0=\max_{a_i\in A} \xi_0(a_i)$.  $\mu_0$ is length of the {\em
longest} of the shortest individual paths and is thus a lower bound on the
minimal makespan. $\mathcal{F}_\xi$ is built on top of the following
understanding about the maximal makespan ($\mu$) of solutions with sum-of-costs
$\xi$. Let $\Delta=\xi-\xi_0$.
\newtheorem{proposition}{Proposition}

\begin{proposition}\label{prop:upperbound}
    If a solution with sum-of-costs $\xi$ exists then its makespan is at most $\mu \leq \mu_0 + \Delta$.
\end{proposition}

{\bf Proof outline~\cite{SurynekFSB16}} Clearly, if there is a solution of cost
$\xi_0$ then its makespan will be no greater than $\mu_0$. But, we want a
solution of cost $\xi$, which is $\xi_0$ plus some $\Delta$. In the worst-case
all the $\Delta$ extra moves belong to the agent with the largest
shortest-path. Thus, the resulting path
of that agent would be $\mu_0+\Delta$, as required. $_\blacksquare$\\

Based on Proposition 1, $\mathcal{F}_\xi$ is constructed by generating a {\em
time expansion graph}~\cite{SurynekFSB16} (denoted TEG) of
$\mu_0+\Delta$ layers. A TEG is a directed acyclic graph (DAG) in which the set
of vertices of the underlaying graph $G$ are duplicated for all time-steps from
0 up to the desired number of layers ($\mu=\mu_0+\Delta$). Possible actions
(move along edges or wait) are represented as directed edges between successive
time steps. Formally a TEG with $\mu$ layers is defined as follows:

\begin{definition}
    \textit{Time expansion graph}  of depth $\mu$ is a digraph $(V,E)$ where
    $V=\{u^t_j|t=0,1,...,\mu \wedge u_j \in V\}$ and $E \subseteq \{
    {(u_j^t,u_k^{t+1})} | t=0,1,...,\mu-1\ \wedge (\{u_j,u_k\} \in E \vee j=k) \}$.
    \label{def:modifiedExpansionGraph}
\end{definition}

\begin{figure}[t]
    \centering
    \includegraphics[trim={6cm 20.9cm 4cm 4.5cm},clip,width=0.55\textwidth]{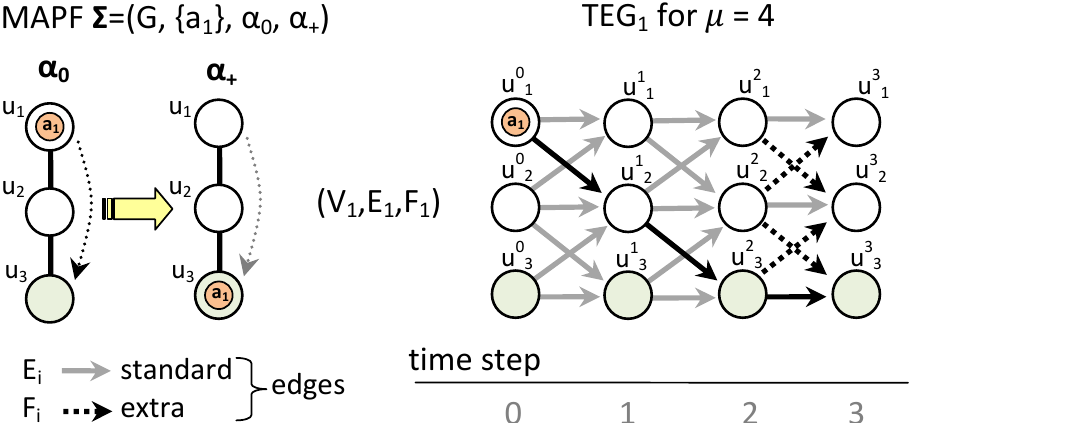}
    \vspace{-0.4cm}\caption{A TEG for an agent that needs to go from $u_1$ to $u_3$.}
    \label{figure-TEGi}
\end{figure}

Figure~\ref{figure-TEGi} illustrates a TEG with 4 layers.
$\mathcal{F}_\xi$ has a a propositional variable
for every pair of agent and edge in the TEG. Setting this variable to \texttt{TRUE}  represents
that the edge is traversed by that agent. Thus, an assignment to these variables represents a solution the the MAPF problem. Appropriate constraints are added to $\mathcal{F}_\xi$ to make sure the solution is valid.

To verify that a solution to $\mathcal{F}_\xi$ represents a solution with sum of costs lower than $\xi$, we add a {\em cardinality constraint} over these agent-edge variables.
Cardinality constraint is a constraint that allows counting variables set to true in a formula and in general bound a numeric cost.
The SAT literature offers several techniques for encoding a \textit{cardinality
    constraint}~\cite{DBLP:conf/cp/BailleuxB03,DBLP:conf/cp/SilvaL07}. Formally, for a
bound $\lambda \in \mathbb{N}$ and a set of propositional variables
$X=\{x_1,x_2,...,x_k\}$ the \textit{cardinality constraint}
$\leq_{\lambda}{\{x_1,x_2,...,x_k\}}$ is satisfied iff the number of variables
from the set $X$ that are set to \texttt{TRUE} is $ \leq \lambda$.
Actually we use the cardinality constraint to bound the number of edges each agent traverses {\em in addition} to the first $\xi_0(a_i)$ edges. So, that the cardinality constraint
in fact ensures that the number of such extra moves is at most $\Delta$. This is done by modifying the TEG, marking some edges are standard and others as extra (see Figure~\ref{figure-TEGi}). We do so for efficiency reasons, following Surynek et al. (\citeyear{SurynekFSB16}).

\begin{algorithm}[t!]
\begin{footnotesize}
\SetAlgoLined \SetKwBlock{NRICL}{\textsc{Mdd-Sat}(MAPF
$\Sigma=(G=(V,E),A,\alpha_0,\alpha_+)$)}{end} \NRICL{
   $\mu_0=\max_{a_i\in A} \xi_0(a_i)$\\
   $\Delta \gets 0$\\
   \While {Solution not found} {
         $\mu=\mu_0 + \Delta$ \nllabel{line:setMu}\\
         $\xi=\xi_0 +\Delta$ \nllabel{line:setXi}\\
         $\mathcal{F}_{\xi} \gets$ Build-Formula($\Sigma$, $\mu$, $\Delta$)\\
         Solution $\gets$ Run-SAT-Solver($\mathcal{F}_{\xi}$)\\
        \lIf {Solution not found}{$\Delta \gets \Delta+1$ \nllabel{line:increaseDelta}}
    }
    \Return Solution\\
} \caption{Sum-of-costs optimal SAT-based solving} \label{alg-MDD-SAT}
\end{footnotesize}
\end{algorithm}

Algorithm~\ref{alg-MDD-SAT} summarizes the \textsc{Mdd-Sat} algorithm. $\Delta$ is initialized as zero and in every iteration it is increased  (Line~\ref{line:increaseDelta}). $\mu$
is set to $\mu_0+\Delta$ (Line~\ref{line:setMu}) and $\xi$ to $x_0+\Delta$ (Line~\ref{line:setXi}). Next the formula $\mathcal{F}_\xi$ is built,
representing the decision problem of asking
whether there is a solution with sum-of-costs $\xi$
and makespan $\mu$. A SAT solver is tasked to check if $\mathcal{F}_\xi$ is solvable.
If such a solution exists, it is returned. Otherwise $\Delta$ and consequently $\xi$
and $\mu$ are incremented by 1 and another iteration of building $\mathcal{F}_\xi$ and running the SAT solver is activated. This algorithm is complete
but cannot detect unsolvability. However, this can be
detected in polynomial time using other algorithms~\cite{kornhauser1984coordinating}.



\section{From Optimal to Suboptimal Solver}
To convert SAT-MDD (Algorithm~\ref{alg-MDD-SAT}) to a suboptimal any solution algorithm, we simply remove the cardinality constraints from the construction of $\mathcal{F}_\xi$. 
Let $\mathcal{F}$ denote the resulting formula. 
Since $\mathcal{F}$ has all the constraints in $\mathcal{F}_\xi$ except the cardinality constraints,
then clearly a satisfying assignment to $\mathcal{F}$ still represents a feasible solution (no collisions between agents etc.). Since $\mathcal{F}$ is less constrained than $\mathcal{F}_\xi$, we expect it to be solved faster.
Indeed, we observed this in our preliminary experiments.
Using $\mathcal{F}$ in Algorithm~\ref{alg-MDD-SAT} instead of $\mathcal{F}_\xi$ looses, however sum-of-cost optimality.

Hence, replacing $\mathcal{F}_\xi$ with $\mathcal{F}$ in Algorithm~\ref{alg-MDD-SAT} leads to a sub-optimal version of the \textsc{Mdd-Sat} solver that is faster than the optimal version. 
We refer to this unbounded version of \textsc{Mdd-Sat} as {\bf u\textsc{Mdd-Sat}}. 
A key question is what is the suboptimality of the solutions u\textsc{Mdd-Sat} returns? Is it really unbounded?
We show later that even without the cardinality constraints, the
suboptimality of the solutions outputted is bounded, due to how $\mathcal{F}$ is constructed. 
Next, we show how to control the suboptimality of the returned solution by introducing a relaxed version of the  optimal cardinality constraints, allowing the algorithm's user to balance runtime and suboptimality. 

\subsection{Bounded Suboptimal SAT-based Solver}

The key to our bounded-suboptimal SAT-based solver is
that it modified the $\Delta$ parameter
used in construction of $\mathcal{F}_\xi$.
In SAT-MDD, $\Delta$ is incremented by one in every iterations.
Allowing $\Delta$ parameter to be less restrictive; that is, replace $\Delta$
with $\Delta'=\Delta+\delta$, where $\delta \geq 0$ is an
integer value, produces formula of the same size but representing more
solutions.\footnote{The change from $\Delta$ to $\Delta'$ does not affect the number of clauses that represent the cardinality constraint, because we coded the cardinality constraints using a sequential counter, whose size is proportional to the number of
propositional variable involved but not to the value of the bound \cite{DBLP:conf/cp/Sinz05}.}
Since $\Delta'>\Delta$, we expect
a formula with the sum-of-costs bounded by $\Delta'$ to be easier to
solve than that with the original $\Delta$.

The following proposition shows that for a solvable MAPF $\Sigma$ the sum-of-costs
of the solution obtained by the above process differs from the optimal one by
at most $\delta$. Let us denote the formula $\mathcal{F}_\xi$ constructed for a TIG with $\mu$ layers (representing a makespan of $\mu$) and $\Delta$ parameter as $\mathcal{F}(\mu,\Delta)$.

\begin{proposition}
Let $\delta$ be a non-negative integer and
let $\mathcal{F}(\mu_0+\Delta,\Delta+\delta)$ be the first satisfiable formula
encountered in the sequence of formulae $\mathcal{F}(\mu_0,\delta)$,
$\mathcal{F}(\mu_0+1,1+\delta)$,...,$\mathcal{F}(\mu_0+\Delta-1,\Delta+\delta-1)$,
$\mathcal{F}(\mu_0+\Delta,\Delta+\delta)$. Then solution represented by
$\mathcal{F}(\mu_0+\Delta,\Delta+\delta)$ has sum-of-costs $\xi \leq \xi_{opt} +
\delta$ where $\xi_{opt}$ is the optimal sum-of-costs for $\Sigma$.
\label{prop:bound}
\end{proposition}

\noindent {\bf Proof:} Formula $\mathcal{F}(\mu_0+\Delta-1,\Delta+\delta-1)$ in the penultimate iteration was not solvable. This means that no solution of makespan at most $\mu_0+\Delta-1$ and sum-of-costs at most $\xi_0+\Delta+\delta - 1$ exists. But we also know that all solutions of sum-of-costs $\xi_0+\Delta-1$ fit under the makespan of at most $\mu_0+\Delta-1$. Hence unsolvability of formula $\mathcal{F}(\mu_0+\Delta-1,\Delta+\delta-1)$ together with $\delta \geq 0$ implies that there is no solution of sum-of-costs $\xi_0+\Delta-1$ at all. Therefore, the optimal sum-of-costs is at least $\xi_0+\Delta$. The solvability of $\mathcal{F}(\mu_0+\Delta,\Delta+\delta)$ tells that there is a solution of $\Sigma$ of sum-of-costs  $\xi_0+\Delta+\delta$ which differs from the optimum by at most $\delta$. $_\blacksquare$\\

Observe that the only property of $\delta$ we used was that it is a
non-negative integer but there is no requirement that it must be constant
across individual iterations of the algorithm. Proposition~\ref{prop:bound} holds
even if we use a non-negative $\delta$ as a function of $\Delta$ instead of a
constant. This property can be used to modify the above SAT-based framework to
an $(1+\epsilon)$-bounded suboptimal algorithm. %

\begin{corollary}
Given an error $\epsilon>0$ the iterative SAT-based suboptimal framework can
modified to an $(1+\epsilon)$-bounded suboptimal algorithm by appropriate setting of
$\delta(\Delta)$.
\end{corollary}

\noindent {\bf Proof:} Let $\delta(\Delta)=\epsilon \cdot (\xi_0+\Delta)$. Hence the sum-of-costs of the solution returned by the algorithm is at most $(1+\epsilon) \cdot (\xi_0+\Delta)$ while the optimum is at least $\xi_0+\Delta$ hence the ratio between the sum-of-costs of returned solution and the sum-of-costs of the optimal one is at most $(1+\epsilon)$. $_\blacksquare$\\

The pseudo-code of the $(1+\epsilon)$-bounded suboptimal SAT-based algorithm is
presented as Algorithm \ref{alg-eMDD-SAT}. We refer to this algorithm as {\bf e\textsc{Mdd-Sat}}.


\begin{algorithm}[t]
\begin{footnotesize}
\SetAlgoLined \SetKwBlock{NRICL}{e\textsc{Mdd-Sat}(MAPF
$\Sigma=(G=(V,E),A,\alpha_0,\alpha_+)$,$\epsilon$)}{end} \NRICL{
   $\mu_0=\max_{a_i\in A} \xi_0(a_i)$\\
   $\xi_0 = \sum_{a_i\in A} \xi_0(a_i)$ \\
   $\Delta \gets 0$\\
   \While {Solution not found} {
         $\mu=\mu_0 + \Delta$\\
         $\Delta' \gets \Delta + \epsilon \cdot (\xi_0 +\Delta)$ \\
         $\mathcal{F}(\mu,\Delta') \gets$ Build-Formula($\Sigma$, $\mu$, $\Delta'$)\\
         Solution $\gets$ Consult-SAT-Solver($\mathcal{F}(\mu,\Delta')$)\\
        \If {Solution not found}{
          $\Delta \gets \Delta+1$\\
        }
    }
   \Return Solution\\
} \caption{e\textsc{Mdd-Sat}, an $(1+\epsilon)$-bounded suboptimal SAT-based MAFP solver} \label{alg-eMDD-SAT}
\end{footnotesize}
\end{algorithm}

Note that a further minor improvement of the pseudo-code could be done which
exploits the original optimization of the formula. Observe that in any solution 
to a MAPF problem it holds that $\mu \leq \xi
\leq {m\cdot\mu}$. Therefore, 
if $\xi_0+\Delta+\delta(\Delta) \geq \mu \cdot k$.
then there is no need to add cardinality any constraints to 
$\mathcal{F}_\xi$, as the solution is guaranteed to be bounded by $\mu \cdot k$.

This inequality represents a limit of the degree of
relaxation achievable by allowing more freedom over the cost bound imposed by
the cardinality constraint. Hence the $(1+\epsilon)$-bounded suboptimal
SAT-based algorithm tends to be near optimal anyway. Precisely, efectively the
algorithm will be $\big(\frac{k \cdot (\mu_0 +
\Delta)}{\xi_0+\Delta}\big)$-bounded in the worst case.

\section{Experimental Evaluation}

We performed a large set of experiments to evaluate
u\textsc{Mdd-Sat} and e\textsc{Mdd-Sat}, our suggested any solution 
and bounded suboptimal versions of \textsc{Mdd-Sat}. We used various
4-connected grids as the underlying graphs.

\begin{figure}[h]
\centering
\includegraphics[trim={2.5cm 22.5cm 6cm 2.7cm},clip,width=0.55\textwidth]{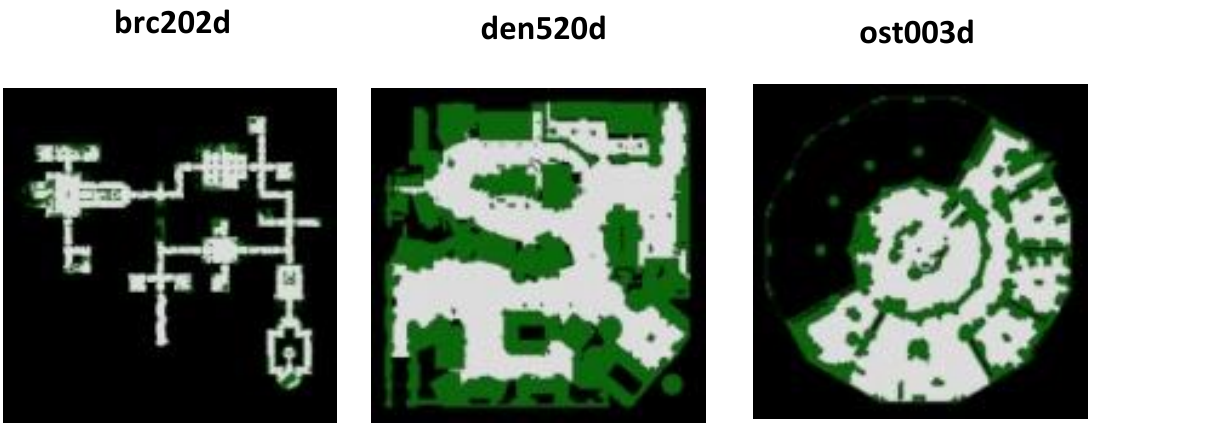}
\vspace{-0.4cm}\caption{Dragon Age maps  include: narrow corridors in \texttt{brc202d}, large open space in \texttt{den520d}, and open space with almost isolated rooms in \texttt{ost003d}.} \label{figure-maps}
\end{figure}

(i) The first set of small densely populated instances consisted of girds of
sizes \texttt{8$\times$8}, \texttt{16$\times$16}, and \texttt{32$\times$32}
with $10\%$ nodes occupied by obstacles. To obtain instances of various
difficulties the number of agents was varied from 1 to 32, 1 to 128, and 1 to
256 in case of \texttt{8$\times$8}, \texttt{16$\times$16}, and
\texttt{32$\times$32} grids respectively (the step was varied from 1 in the
range of small units of agents to 16 in the range of hundreds of agents). Ten
random  instances were genereted for each number of agents by randomly choosing an initial position
and then performing a random walk to set the target position. 

(ii) Instances of the second testing set are based on three structurally
different large maps taken from Sturtevant's
repository~\cite{sturtevant2012benchmarks}. These are Dragon Age Origion (DAO)
maps denoted as \texttt{brc202d}, \texttt{den520d}, and \texttt{ost003d} which
are a standard benchmark for MAPF (see Figure \ref{figure-maps}). Again the
number of agents was varied from 1 to 256 to obtain instances of various
difficulties (the step ranged from 1 to 16) and 10 random instances were
generated for each number of agents.

All tests were run on a machine with CPU Intel i7 3.2 Ghz, 8 GB RAM under
Ubuntu Linux 15 and Windows 10 respectively. The timeout for all solvers has
been set to 500 seconds.

\subsection{Evaluation of the Unbounded Case}

\begin{figure}[h]
\includegraphics[trim={3.0cm 20.5cm 3.0cm 2.8cm},clip,width=0.48\textwidth]{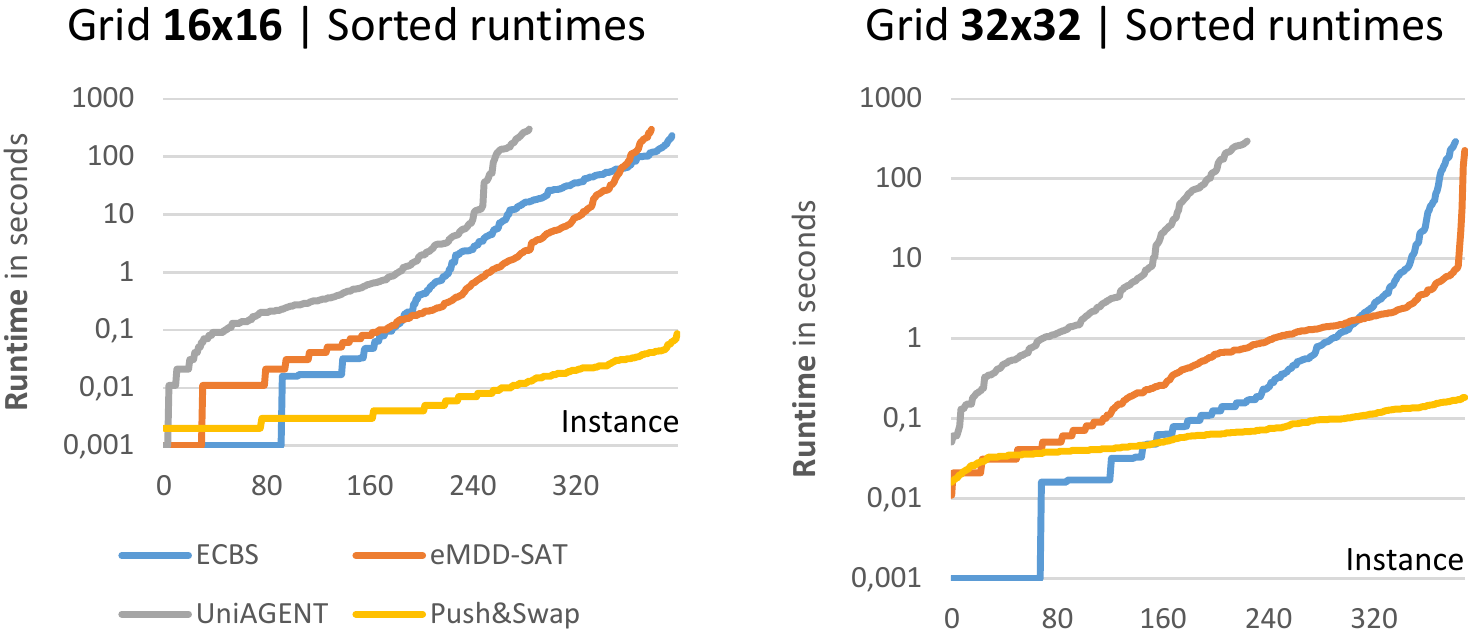}
\caption{Runtimes of unbounded variants on grids of size  \texttt{16$\times$16}, and \texttt{32$\times$32}.}
\label{fig_exp_grids_unb}
\end{figure}

In this section we evaluate the performance of u\textsc{Mdd-Sat}, our any solution suboptimal SAT-based solver. 
We compared u\textsc{Mdd-Sat} with two suboptimal
algorithms that are by design unbounded: 
\textsc{Push-and-Swap} \cite{LunaB11,DBLP:journals/jair/WildeMW14}, which is
a polynomial time rule-based algorithm, 
and \textsc{UniRobot}\cite{DBLP:conf/ijcai/Surynek15}, which is a SAT-based algorithm that reduces MAPF
with $k$ agents to a problem of finding $k$ vertex disjoint paths \cite{SEYMOUR1980293}. 
We also compared u\textsc{Mdd-Sat} against \textsc{Ecbs}~\cite{barer2014suboptimal} a
state-of-the-art bounded-suboptimal algorithm that is based on the CBS MAPF solver. 
To make the comparison with unbounded MAPF solver fair, we set the suboptimality bound of \textsc{Ecbs} to a very large number 
(500).

\begin{figure}[h]
\includegraphics[trim={3.0cm 13.2cm 3.0cm 2.8cm},clip,width=0.48\textwidth]{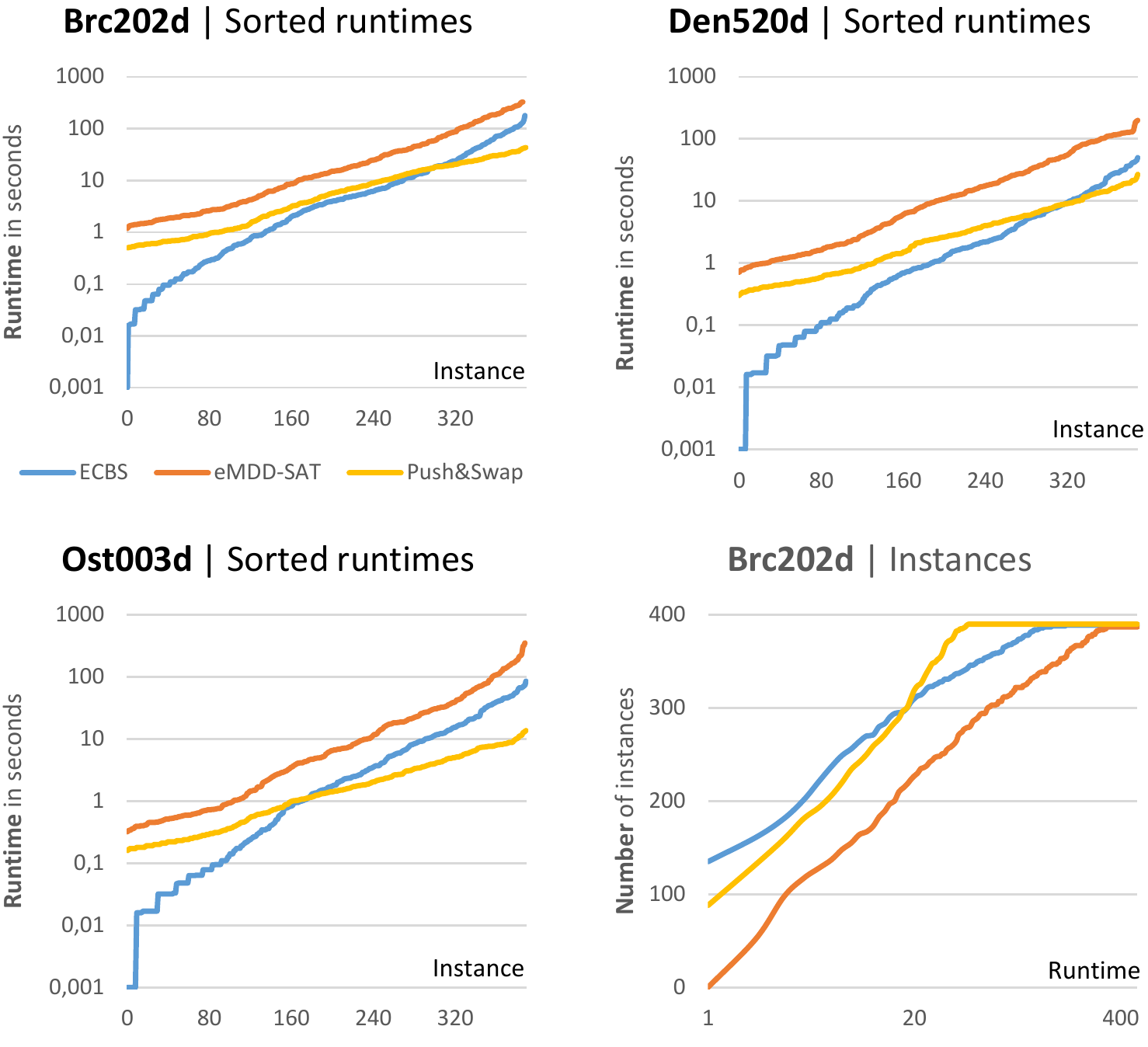}
\caption{Runtimes of unbounded variants on DAO maps \texttt{brc202d}, \texttt{den520d}, and \texttt{ost003d}.}
\label{fig_exp_maps_unb}
\end{figure}

Runtime results of the experiments with the unbounded versions on small grids
and DAO maps are shown in Figures \ref{fig_exp_grids_unb} and
\ref{fig_exp_maps_unb} respectively. Runtimes for all testing instances that
were below the limit of 500 seconds were sorted and shown in the figure (the
x-axis corresponds to ordering of instances according to increasing runtime and
the y-axis corresponds to runtime in seconds). The intuitive understanding of
this presentation is that the faster algorithm has its line in the lower
part of the figure. 

Consider first the runtime results for the 16x16 and 32x32 grids (Figure~\ref{fig_exp_grids_unb}). 
\textsc{Push-and-Swap} is the fastest algorithm in these small grids and \textsc{UniRobot} turned out to be worst performing algorithm. The comparison of \textsc{Ecbs} and e\textsc{Mdd-Sat} shows that in the easier instances (those that are sorted in the left-hand side of the $x$-axis),  \textsc{Ecbs} is faster, while for the harder instnances (those on the right-hand side of the $x$-axis)
e\textsc{Mdd-Sat} performs better. 

\begin{figure}[h]
\includegraphics[trim={3.0cm 13.2cm 3.0cm 2.8cm},clip,width=0.48\textwidth]{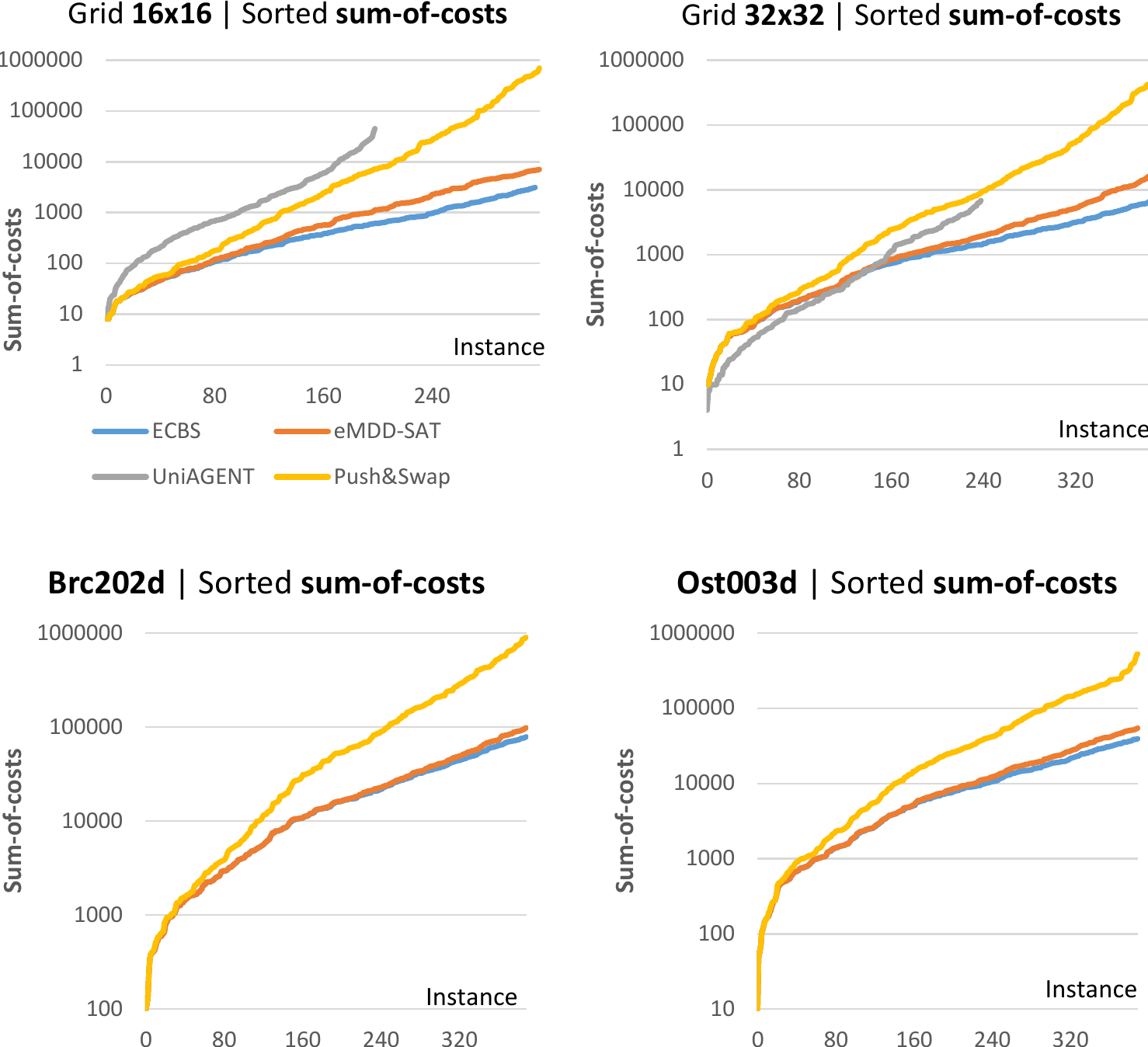}
\caption{Sum-of-costs of unbounded variants on small grids and DAO maps.}
\label{fig_exp_unb_cost}
\end{figure}

Now consider the runtime results on the DAO maps (Figure~\ref{fig_exp_maps_unb}), which are much larger than the aforementioned grids. Here too, \textsc{UniRobot} turned out to be the worst performing algorithm, 
and in general not applicable on large DAO maps. 
The bottom-right plot in Figure~\ref{fig_exp_maps_unb}, 
which shows the number of instances solved by the remaining algorithms as a function of the runtime, 
that is, for a given $x$ value the $y$ value shows the number of instances solved 
given $x$ seconds. All three algorithms (\textsc{Ecbs}, \textsc{Push-and-Swap}, and u\textsc{Mdd-Sat}) managed to solve all instances within the time limit, 
but u\textsc{Mdd-Sat} was somewhat slower than the other two. 
Consider the other plots in Figure~\ref{fig_exp_maps_unb} we can conclude that in this domain
\textsc{Ecbs} was in general faster.

While the compared algorithms do not provide a bound on the sum-of-costs of their solutions, 
it may still be of interest in practice. We observed that the sum-of-costs 
of the solutions found by tested algorithms were significantly different from the optimum and from each other. Figure~\ref{fig_exp_unb_cost} shows 
	the sum-of-costs of the solutions found for the DAO instances.
	This presentation is similar to the plots in the previous figures, 
	but here the instances are sorted according ot their sum-of-costs. 
	The interpretation is the same: lower curves corresponds to finding lower sum-of-costs. 
	The results show that both \textsc{UniRobot} and \textsc{Push-and-Swap} generate
worse solutions than \textsc{Ecbs} and  e\textsc{Mdd-Sat}. The solutions quality returned by \textsc{Ecbs} and  e\textsc{Mdd-Sat} are comparable, with slightly better solutions found by \textsc{Ecbs} in some cases.

Altogether we can conclude that for unbounded suboptimal case 
 u\textsc{Mdd-Sat} is a reasonable option: perhaps not always the fastest or the one with the lowest sum-of-costs, but comparable to the state-of-the-art. This is encouraging, especially since 
 if SAT solvers continue to become better, the performance of SAT-based algorithms such as 
  u\textsc{Mdd-Sat} will continue to improve.



\subsection{Evaluation of the Bounded Case}

Next, we conducted experiments to evaluate e\textsc{Mdd-Sat}, our bounded-suboptimal  \textsc{Mdd-Sat} variant. Here we only compared againts \textsc{Ecbs} as the other algorithms (\textsc{Push-and-Swap} and \textsc{UniRobot}) do not guarantee a bounded-suboptimal solution. 
The first set of experiments evaluate the behavior of both algorithms for different values of $1+\epsilon$, i.e., for different required suboptimality bounds. 
The same set of instances used for the unbounded experiments were also used here. 

\begin{figure}[h]
\includegraphics[trim={2.7cm 20.5cm 2.7cm 2.8cm},clip,width=0.48\textwidth]{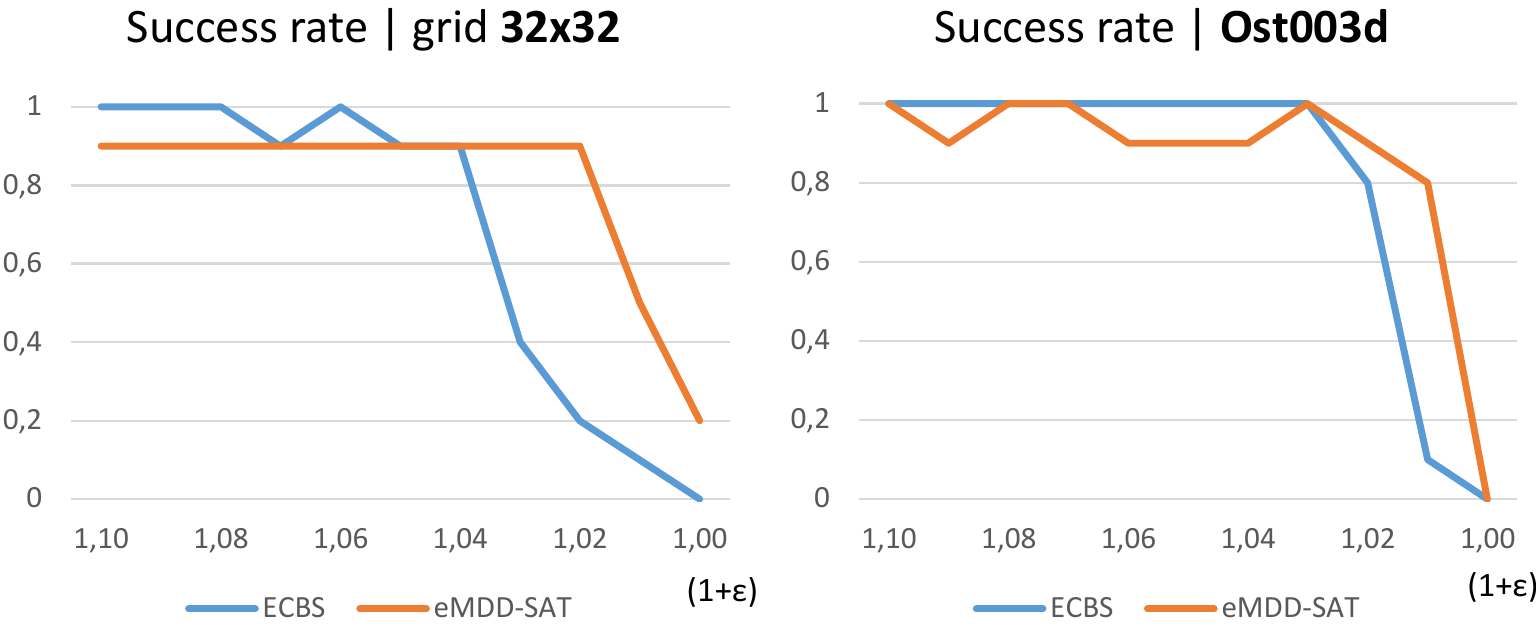}
\caption{Success rate of the bounded variant with $(1+\epsilon)$ ranging from 1.1 to 1.0 (optimal case). Grid\texttt{32$\times$32} contains 100 agents and DAO map \texttt{ost003d} 200 agents.}
\label{fig_exp_succ}
\end{figure}

First, we measured the {\em success rate} of each algorithm, which is the the ratio of successfully solved instances under a predetermined time limit. The time limit in our experiments was 500 seconds. 
Figure~\ref{fig_exp_succ} shows the algorithms' success rate ($y$-axis) as a function of the required suboptimality bound (the $x$-axis), which ranges from 1.1 to 1.0. Results for \texttt{32$\times$32} with 100 agents and  \texttt{ost003d} with 200 agents are shown. It can be observed that  e\textsc{Mdd-Sat} is better than \textsc{Ecbs} for closer to optimal suboptimality bounds, 
outperforming  \textsc{Ecbs} starting at bound $(1+\epsilon)=1.02$ and lower. 
For the 32x32 grids, which are more dense than the DAO map, the advantage of e\textsc{Mdd-Sat}
even begin earlier, again highlighting the advantage of SAT-based algorithms in harder problems. 
Next, we focus our evaluation on bound $(1+\epsilon)=1.01$, to focus on the cases 
where e\textsc{Mdd-Sat} is effective. 

\begin{figure}[h]
\includegraphics[trim={3.0cm 13.2cm 3.0cm 2.8cm},clip,width=0.48\textwidth]{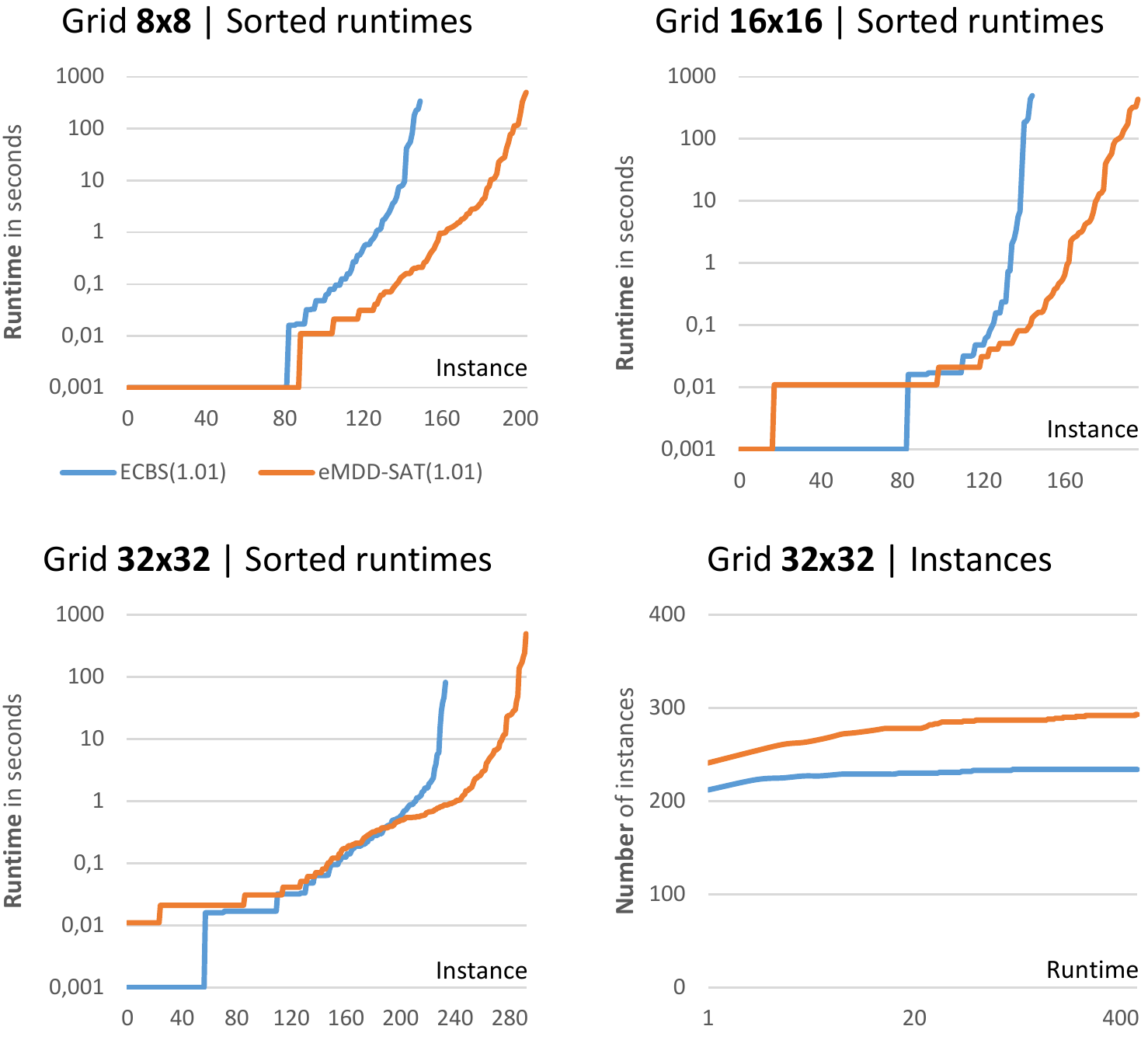}
\caption{Runtimes of bounded variants ($\epsilon=0.01$) on grids of size \texttt{8$\times$8} and \texttt{16$\times$16}, and \texttt{32$\times$32}.}
\label{fig_exp_grids_101}
\end{figure}

\begin{figure}[h]
\includegraphics[trim={3.0cm 13.2cm 3.0cm 2.8cm},clip,width=0.48\textwidth]{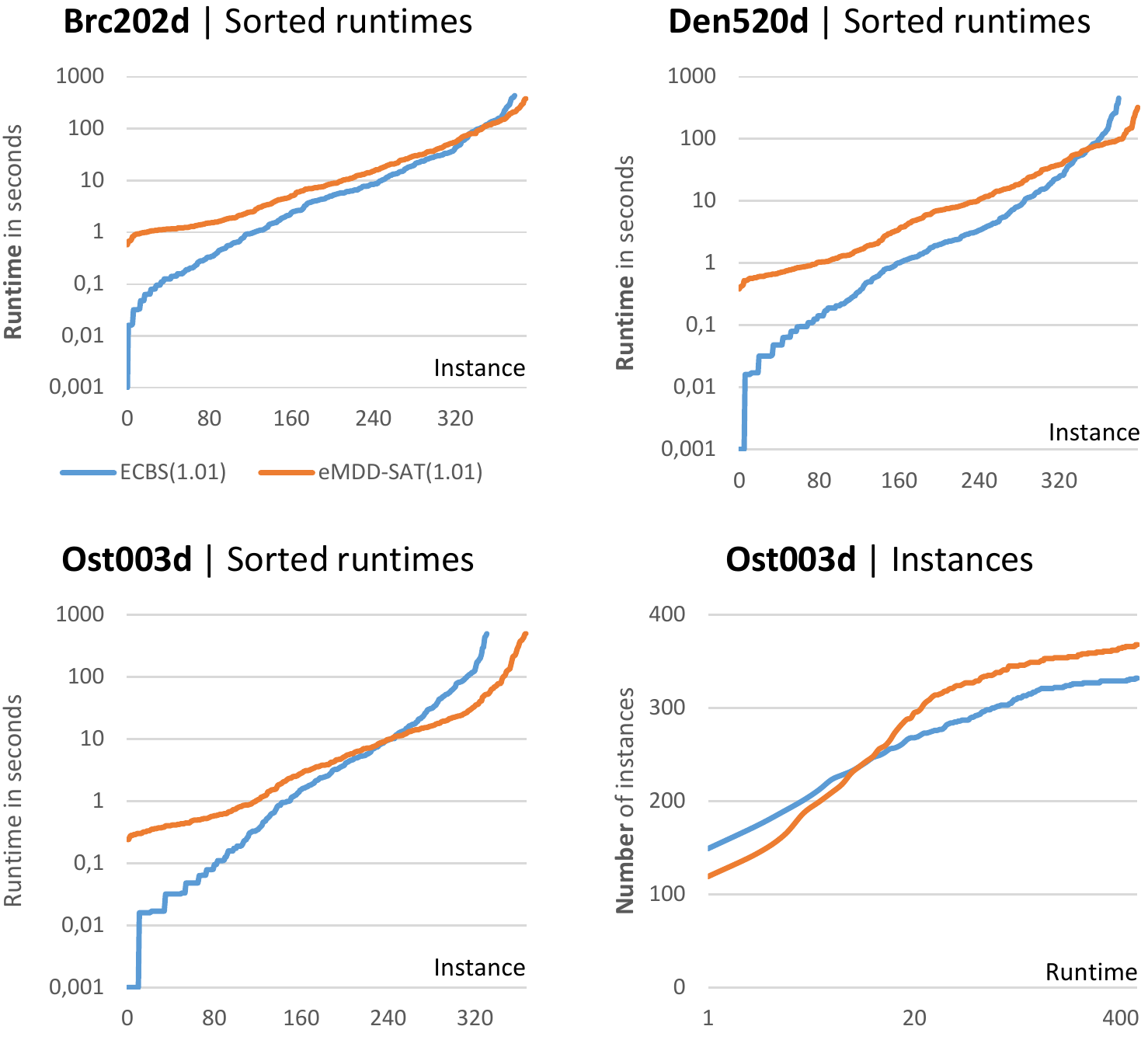}
\caption{Runtimes of bounded variants ($\epsilon=0.01$) on DAO maps \texttt{brc202d}, \texttt{den520d}, and \texttt{ost003d}.}
\label{fig_exp_maps_101}
\end{figure}

Results for $(1+\epsilon)=1.01$  for small girds and DAO maps are presented in figures \ref{fig_exp_grids_101} and \ref{fig_exp_maps_101}. In this we can observe that \textsc{Mdd-Sat} tends to be faster in all small grids for the harder problems. In our analysis (results not shown for space limitation), we observed that these were the cases with higher density of agents. 

Results for DAO maps indicate that in easier instances containing fewer agents \textsc{Ecbs} is faster. However with the increasing difficulty of instances and density of agents the gap in performance is narrowed until e\textsc{Mdd-Sat} starts to perform better in harder instances. This trend is best visible on the \texttt{ost003d} map.

Let us note that the maximum achievable $\epsilon$ by relaxing the cardinality constraint within the suboptimal e\textsc{Mdd-Sat} approach for DAO maps is: $\epsilon=1.47$ for \texttt{brc202d}, $\epsilon=1.33$ for \texttt{den520d}, and $\epsilon=1.12$ for \texttt{ost033d} all cases with 200 agents. Setting these or greater bounds in e\textsc{Mdd-Sat} is equivalent to complete removal of the cardinality constraint. That is, it is equivalent to running u\textsc{Mdd-Sat}.

\section{Conclusions}
The SAT-based approach represented by e\textsc{Mdd-Sat} has an advantage of using the learning mechanism built-in the external SAT solver. On the other hand, search based methods represented by \textsc{Ecbs} are specially designed for solving MAPF and do not bring the overhead of a general purpose SAT solver. We attribute the good performance of the e\textsc{Mdd-Sat} approach to clause learning mechanism. 

This conclusion corresponds with the fact that advantage of e\textsc{Mdd-Sat} appears in harder instances with long runs of the SAT solver where the clause learning mechanism has enough time to prune the search space efficiently. On the other hand the SAT-based approach has an overhead of building formula and communication with the external solver which negativelly affects performance in sparsely occupied instances.

One of possible future research directions is to integrate learning mechanism into specialized MAPF solver which would eliminate the overhead of usage of the external SAT solver. Vertices represented within layers of MDD can be regarded as values of a multi-value decision variables representing positions of agents at individual time steps. Learning mechanism over such finite domain variables would be very similar to {\em nogood recording} known from modern CSP solvers \cite{DBLP:books/daglib/0016622}. Reasoning about MAPF in the context of nogood recording and CSP would open door to higher level constraint propagation than that offered by SAT's unit propagation. 
Lastly, we believe that this work will open the way to developing bounded-suboptimal SAT-based algorithms for other planning problems. 

\section{Acknowledgements}
This paper is supported by a project commissioned by the New Energy and Industrial Technology Development Organization Japan (NEDO) and the joint grant of the Israel Ministry of Science and the Czech Ministry of Education Youth and Sports number 8G15027.

\bibliography{SubOpt-MDD-SAT_arXiv-2017}
\bibliographystyle{named}
\end{document}